\pdfoutput=1
\documentclass[11pt]{article}

\usepackage[review]{emnlp2021}

\usepackage{times}
\usepackage{latexsym}

\usepackage[T1]{fontenc}
\usepackage[utf8]{inputenc}

\usepackage{microtype}

\usepackage{color}

\usepackage{amsmath,amsfonts,bm}

\def\eqref#1{equation~\ref{#1}}
\def\1{\bm{1}}

\def\va{{\bm{a}}}

\def\vc{{\bm{c}}}

\def\vf{{\bm{f}}}
\def\vg{{\bm{g}}}
\def\vh{{\bm{h}}}

\def\vo{{\bm{o}}}
\def\vp{{\bm{p}}}
\def\vq{{\bm{q}}}
\def\vr{{\bm{r}}}

\def\vu{{\bm{u}}}
\def\vv{{\bm{v}}}

\def\vx{{\bm{x}}}

\def\vz{{\bm{z}}}

\def\mH{{\bm{H}}}

\def\mU{{\bm{U}}}

\DeclareMathAlphabet{\mathsfit}{\encodingdefault}{\sfdefault}{m}{sl}
\SetMathAlphabet{\mathsfit}{bold}{\encodingdefault}{\sfdefault}{bx}{n}

\def\gA{{\mathcal{A}}}

\def\gD{{\mathcal{D}}}

\def\gL{{\mathcal{L}}}

\def\gO{{\mathcal{O}}}

\def\gQ{{\mathcal{Q}}}

\def\gS{{\mathcal{S}}}

\newcommand{\R}{\mathbb{R}}

\newcommand{\softmax}{\mathrm{softmax}}

\usepackage{multirow}
\usepackage{amsmath}
\usepackage{graphicx}
\usepackage{subfigure}
\usepackage{bbding}
\usepackage{mathtools}
\usepackage{bbding}

\DeclareMathOperator*{\transformerenc}{Transformer-Enc}

\DeclareMathOperator*{\mlp}{MLP}
\DeclareMathOperator*{\attnpool}{Attn-Pool}

\DeclareMathOperator*{\sentimodel}{Senti-Model}
\DeclareMathOperator*{\kl}{KL}
\DeclareMathOperator*{\sigmoidii}{Sigmoid}

\title{Eliminating Sentiment Bias for Aspect-Level Sentiment Classification with Unsupervised Opinion Extraction}

\author{
	Bo Wang\textsuperscript{1}, Tao Shen\textsuperscript{2}, Guodong Long\textsuperscript{2}, Tianyi Zhou\textsuperscript{3,4}, Yi Chang\textsuperscript{1,5,\Envelope} \\
	\textsuperscript{1}School of Artificial Intelligence, Jilin University\\ 
	\textsuperscript{2}Australian AI Institute, School of CS, FEIT, University of Technology Sydney \\
	\textsuperscript{3}University of Washington, Seattle; \textsuperscript{4}University of Maryland, College Park \\
	\textsuperscript{5}International Center of Future Science, Jilin University\\
	\texttt{bowang19@mails.jlu.edu.cn,\{tao.shen,guodong.long\}@uts.edu.au, } \\
	\texttt{tianyizh@uw.edu,yichang@jlu.edu.cn} 
}

\begin{document}
	\maketitle
	
	\begin{abstract}
		Aspect-level sentiment classification (ALSC) aims at identifying the sentiment polarity of a specified aspect in a sentence. ALSC is a practical setting in aspect-based sentiment analysis due to no opinion term labeling needed, but it fails to interpret why a sentiment polarity is derived for the aspect. To address this problem, recent works fine-tune pre-trained Transformer encoders for ALSC to extract an aspect-centric dependency tree that can locate the opinion words. However, the induced opinion words only provide an intuitive cue far below human-level interpretability. Besides, the pre-trained encoder tends to internalize an aspect's intrinsic sentiment, causing sentiment bias and thus affecting model performance. In this paper, we propose a span-based anti-bias aspect representation learning framework. It first eliminates the sentiment bias in the aspect embedding by adversarial learning against aspects' prior sentiment. Then, it aligns the distilled opinion candidates with the aspect by span-based dependency modeling to highlight the interpretable opinion terms. Our method achieves new state-of-the-art performance on five benchmarks, with the capability of unsupervised opinion extraction. 
	\end{abstract}

	\section{Introduction} \label{sec:intro}
	
	Aspect-based sentiment analysis (ABSA) \citep{jiang2011twitter_ABSA} aims to determine sentiment polarity w.r.t. a specified aspect term in a piece of text. 
	For example, in ``\textit{The food is tasty, but the service is terrible}'', the sentiment towards aspect term (AT) ``\textit{food}'' is positive according to the opinion term (OT) ``\textit{tasty}'', while the sentiment towards ``\textit{service}'' is negative according to ``\textit{terrible}''. 
	The most comprehensive setting of ABSA is aspect sentiment triplet extraction (ASTE) \citep{Peng2020Knowing} consisting of a series of subtasks, i.e., aspect extraction, aspect-level sentiment classification and opinion extraction. Thereby, given a piece of text, ASTE can produce a set of triples, i.e., (\textit{Aspect Term}, \textit{Sentiment}, \textit{Opinion Term}), to describe sentiment with details of What, How and Why, so it enjoys full interpretability. 
	Continuing the above example, ASTE can generate triples like (\textit{food}, \textit{Positive}, \textit{tasty}). 
	However, the human annotation on opinion terms is much more label-intensive than traditional sentiment analysis task.

	Therefore, by following many recent works, we target the practical subtask of ASTE, called aspect-level sentiment classification (ALSC). 
	It predicts a three-categorical sentiment (i.e., \textit{positive}, \textit{neutral} or \textit{negative}) of a given aspect term in a sentence. 
	Most recent works capture the modification relation between aspect and opinion terms in an implicit manner, which is usually achieved by integrating graph neural networks (GNNs) over dependency parsing tree into text representation learning \citep{Zhang2019ASGCN,Sun2019CDT,Tang2020DGEDT,wang2020RGAT}. 
	Further, the performance can be significantly boosted when incorporating pre-trained Transformer encoder, e.g., BERT \citep{BERT} in a fine-tuning paradigm \citep{sun2019bert_qa,Tang2020DGEDT,wang2020RGAT,chen2020kumaGCN}. 
	Although these works achieve excellent results even close to humans, they cannot derive the interpretability to explain \textit{why} an aspect is associated with the polarity prediction.

	Luckily, pre-trained Transformer encoder can also be used to explain linguistic knowledge underlying the given text via dependency probing \citep{Clark2019BERT-analysis,Wu2020Perturbed-masking}. This has been exploited by \citet{Dai2021does-syntax} to reveal that, after fine-tuning the pre-trained encoder on ALSC, an aspect-centric dependency tree can be induced to highlight the modifier of an aspect. 
	Intuitively, the highlighted modifier is viewed as an opinion word of the corresponding aspect, which thus, to some extent, brings the interpretability back. 
	
	Nevertheless, compared with the span-formatted opinion terms in ASTE, the opinion word illustrated by the induced dependency tree can only provide an intuitive, noisy, word-level sentiment cue but is far from the human-level interpretability as in opinion extraction. 
	What's worse, as verified by \citet{Huang2020sentiment-bias}, the Transformer pre-trained on large-scale raw corpora tends to internalize terms' intrinsic attributes, so it causes the problem of \textit{sentiment bias} when generating text given a particular term prompt. 
	In ALSC scenario, we found that \textit{sentiment bias} also exists and affects a model to determine the sentiment of an aspect term regardless of its contextual information (e.g., opinion terms). 
	The bias is especially obvious for the aspects that can imply strong sentiment themselves. 
	For example, for ``\textit{There's candlelight and music}'', a model based on pre-trained Transformer is likely to mis-classify the sentiment towards ``\textit{music}'' as positive, whereas the oracle label is neutral.

	In this work, based on the pre-trained Transformer encoder to ensure state-of-the-art ALSC performance, we aim to eliminate sentiment bias in the ALSC scenario while equip the model with human-level aspect-opinion interpretability. 
	
	To this end, we propose a Span-based Anti-bias aspect Representation Learning (SARL) framework for ALSC with unsupervised opinion extraction. 
	First, instead of widely feeding a concatenation of sentence and aspect into a pre-trained Transformer, we adopt a span-level paradigm \citep{hu2019SpanABSA,Zhao2020SpanMlt} which focuses on deriving span representation of an aspect term. 
	Then, we propose an anti-bias aspect encoding module to eliminate the sentiment bias existing in aspect representations, which is achieved by an adversarial learning against the aspect's prior sentiment indicated in SentiWordNet \citep{Esuli2006SentiWordNet}. 
	Next, built upon sentiment-agnostic aspect representation from the above encoder, we propose an aspect-opinion dependency alignment module to capture explicit modifications from opinion term candidates to the targeted aspect, and integrate the modifications into aspect representation via gating. The integrated representation is lastly passed into a neural classifier for sentiment prediction.  
	
	For pairwise aspect-opinion alignment via a span-based model, our work share a high-level inspiration with SpanMlt \citep{Zhao2020SpanMlt}
	but differs in that, SpanMlt targets fully-supervised aspect and opinion terms extraction whereas ours leverages the alignment to empower aspect-opinion interpretability even without opinion supervisions. 
	
	Our main contributions are as follows:
	\begin{itemize}
		\vspace{-3mm}
		\item We propose a span-based aspect encoding module to alleviate the sentiment bias problem from which pre-trained Transformers suffer. 
		\vspace{-3mm}
		\item By a dependency aligner, our model can derive human-level aspect-opinion interpretability w/o opinion extraction supervisions. 
		\vspace{-3mm}
		\item We achieve new state-of-the-art results on 5 ALSC datasets with extensive analyses and present metrics to measure the interpretability. 
	\end{itemize}
	
	\section{Methodology} % [Figure] Model Architecture

\begin{figure*}
		\centering
		\includegraphics[width=0.8\linewidth]{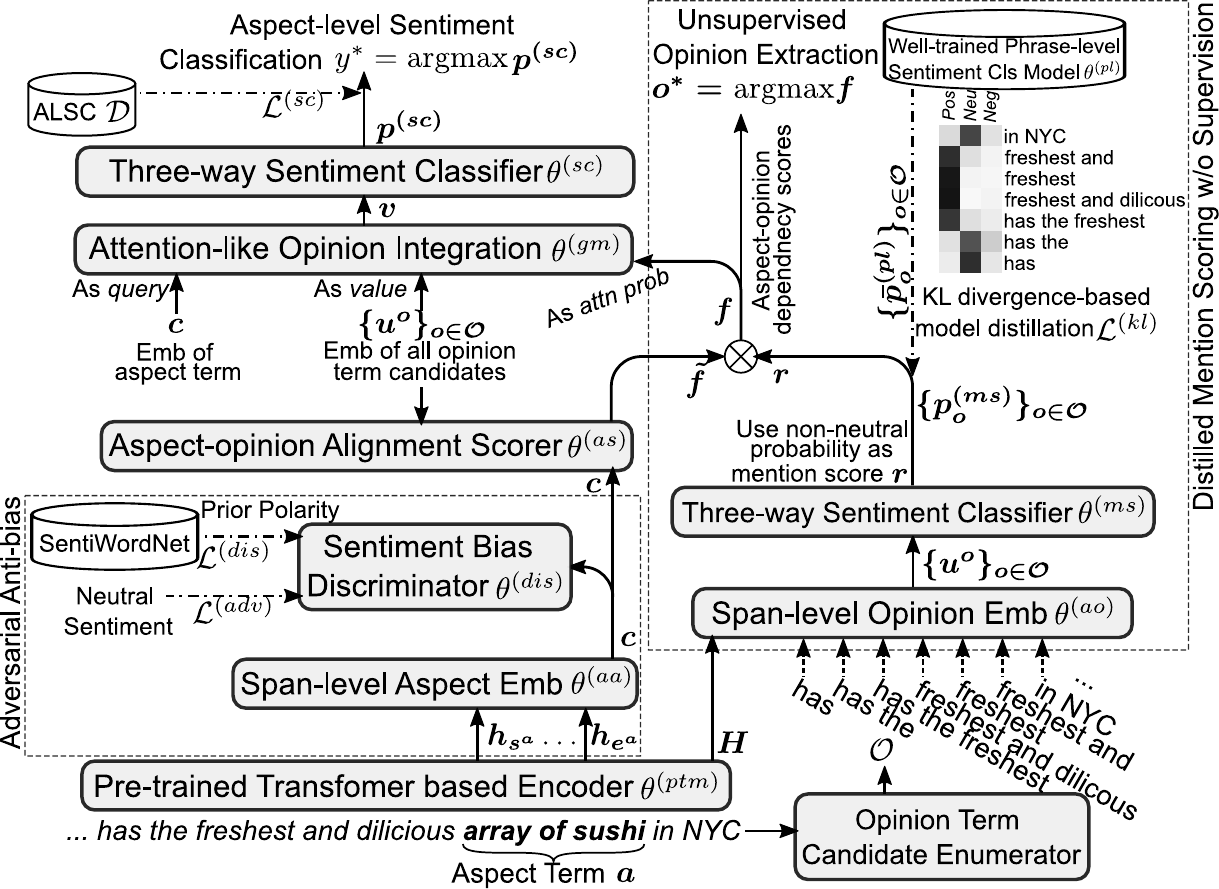}
		\caption{\small An overview of proposed Span-based Anti-bias aspect Representation Learning (SARL) framework. }
		\label{fig:model_architecture}
	\end{figure*}

	This section begins with a task definition of ALSC.
	Then, we present our Span-based Anti-bias aspect Representation Learning (SARL) framework (as in Figure~\ref{fig:model_architecture}) consisting of an adversarial anti-bias aspect encoder (\S\ref{sec:adv_encoder}) and a distilled aspect-opinion dependency aligner (\S\ref{sec:asp-op_dep}). 
	Lastly, we detail the training and inference of the proposed model (\S\ref{sec:train-infer}), including unsupervised opinion extraction. 
	
	\paragraph{Task Definition. }
	Given a sentence with $n$ words, $\vx = [x_1, \dots, x_n]$, ALSC aims to predict a three-categorical sentiment from $\{Positive$, $Neutral$, $Negative\}$ of an aspect term, $a$, where $a$ is a span of $\vx$, from $s^a$ and to $e^a$, i.e., $\va=x_{s^a:e^a}$.

	\subsection{Adversarial Anti-bias Aspect Encoder} \label{sec:adv_encoder}
	
	As suggested by \citet{BERT} and verified by many works, a common practice to tackle pair inputs is feeding the pre-trained Transformer with a concatenation of them. 
	Hence, many ALSC works \citep{Jiang2020METNet,Hou2021graph_emsemble} feed a concatenation of the sentence $\vx$ and an aspect $\va$ into the encoder. 
	Nonetheless, considering that every aspect is a word span of the sentence, it is natural to employ a span-level paradigm for aspect representation, which has been proven as effective as the concatenation models \citep{Zhou2019Span-based_AOC,Zhao2020SpanMlt}.
	Also, a span-level model is more efficient as it predicts the sentiment for all aspect terms in a single feed-forward process. 
	
	\paragraph{Span-level Aspect Representation.}
	We thereby adopt the span-level paradigm to generate aspect representation. Formally, a Transformer is fed with a sentence without additional aspect terms,
	\begin{align}
		\mH = \transformerenc(\vx;\theta^{(ptm)})  \label{eq:sentence_embedding}
	\end{align}
	where, 
	$\mH = [\vh_1, \dots, \vh_n]\in\R^{d\times n}$ denotes contextualized representations corresponding to all the words. 
	Then, given span $[s^a,e^a]$ of each aspect term $\va$, we derive its representation by
	\begin{align}
		\vc^a \!=\! [\vh_s^a; \!\vh_e^a; \!\attnpool(\mH_{s^a:e^a} ;\!\theta^{(aa)})],
		\label{eq:span_asp_emb}
	\end{align}
	where $[;]$ is vector concatenation, $\attnpool(\cdot)$ is attention pooling to generate sequence-level embedding with multi-layer perceptron (MLP)-derived weights \citep{Lin2017self-attentive,Liu2016inner-attention}.
	And $\vc^a\in\R^{3d}$ is the resulting span-level aspect representation. 
	In the remainder, we omit the aspect indicator, $a$,  if no confusion caused. 
	But, as briefed in \S\ref{sec:intro}, sentiment bias will occur since the Transformer encoders pre-trained on large-scale text corpora incline to internalize the sentiment polarity of aspect terms. 
	Hence, directly applying a classifier to the aspect representation $\vc$ for its sentiment prediction will be affected. 
	
	\paragraph{Adversarial Anti-Bias Module.} 
	A promising way is leveraging adversarial learning to eliminate the sentiment bias of a particular aspect. Open questions still remain about how to define the \textit{sentiment bias} as discriminator's objective and how to fool the discriminator for our anti-bias purpose. As an answer to the first answer, we resort to external sentiment knowledge, SentiWordNet \citep{Esuli2006SentiWordNet}, for its prior three-categorical sentiment for common lexicons and phrases. 
	Given an aspect $\va$ from a sentence, we can easily obtain its prior sentiment polarity $y^{(pr)}$ by querying SentiWordNet. 
	The prior sentiment polarity is the intrinsic attribute of an aspect, which thus can be viewed as its sentiment bias. 
	Hence, we train a sentiment bias discriminator towards $y^{(pr)}$ to exploit bias information underlying the aspect representation. Formally, we present an MLP-based neural classifier upon $\vc$ as the discriminator, i.e., 
	\begin{align}
		\vp^{(pr)} = \softmax(\mlp(\vc^a; \theta^{(dis)})) \in\R^{3}.
	\end{align}
	Next, training loss of this discriminator is
	\begin{align}
		\gL^{(dis)}_{\theta^{(dis)}} = - \sum\nolimits_\gD \sum\nolimits_{\gA} \log \vp^{(pr)}_{[\hat y = y^{(pr)}]}, \label{eq:discriminator_loss}
	\end{align}
	where $\gD$ denotes ALSC dataset with sentence-level samples, where $\gA$ denotes all aspects in a sentence, and $\vp^{(pr)}_{[\hat y = y^{(pr)}]}$ denotes fetching the probability value corresponding to prior sentiment $y^{(pr)}$. 
	
	On the other side, to eliminate sentiment bias, the span-level aspect encoder aims to fool the discriminator towards generating neutral-sentiment representation. The adversarial loss is written as
	\begin{align}
		\gL^{(adv)}_{\theta^{(ptm)}, \theta^{(aa)}} = - \sum\nolimits_\gD \sum\nolimits_{\gA} \log \vp^{(pr)}_{[\hat y = \text{Neutral}]}. \label{eq:adversarial_loss}
	\end{align}
	With the adversarial learning between $\gL^{(dis)}$ and $\gL^{(adv)}$, the aspect representation $\vc$ can escape from sentiment bias and become sentiment-agnostic. 
	
	\subsection{Aspect-Opinion Dependency Aligner} \label{sec:asp-op_dep}
	
	As opinion term is also a span of words, it is intuitive to use a span-based alignment model \citep{Lee2017e2e_corefernce} for explicit modification from opinion terms to the aspect. 
	This model is first proposed for co-reference resolution to align an entity mention with its antecedent, but usually requires full supervisions.  
	For example, it has been adapted to extract pair-wise aspect and opinion terms \citep{Zhao2020SpanMlt} in ABSA, and the supervisions include both opinion terms and aspect-opinion associations but none of them is available in ALSC. 
	
	Fortunately, built upon the sentiment-agnostic aspect representation derived from the above encoder, the span-based alignment model is likely to automatically learn the dependency between opinion and aspect spans, which is guided by the aspect's sentiment label. To this end, we present an aspect-opinion dependency aligner w/o supervisions.

	\paragraph{Opinion Term Candidates.}
	Similar with \citet{Lee2017e2e_corefernce} and \citet{Zhao2020SpanMlt}, we first enumerate all the possible spans as candidates of opinion terms. 
	Given the sentence $\vx = [x_1, \dots, x_n]$, an opinion term candidate $\vo$ is a span of words from $s^o$ and to $e^o$, i.e., $\vo = x_{s^o:e^o}$. A set of all the opinion term candidates can be written as
	\begin{align}
		\notag \gO = \{x_{s^o:e^o}| &1 \le s^o \le e^o \le n \wedge e^o - s^o \le l  \\
		&\wedge ( e^o < s^a \vee s^o > e^a ) \},
	\end{align}
	where the last condition means no overlap between an arbitrary oracle aspect term and its opinion candidates, and $m \coloneqq |\gO|$. Then, we can easily derive the representation of each candidate $\vo \in \gO$ given the contextualized embedding $\mH$ by the encoder in Eq.(\ref{eq:sentence_embedding}). We adopt the same scheme to generate span-level opinion embedding as in Eq.(\ref{eq:span_asp_emb}), i.e., 
	\begin{align}
		\vu^o \!=\! [\vh_{s^o}; \!\vh_{e^o}; \!\attnpool(\mH_{s^o:e^o} ;\!\theta^{(ao)})], % \!\in\!\R^{3d}, 
		\label{eq:span_embedding}
	\end{align}
	where the resulting vector $\vu^o\in\R^{3d}$ represents the candidate $\vo$. Thus, we can get a set of candidate representations, $\mU = \{\vu^o\}_{\forall o\in\gO} \in\R^{3d\times m}$.

	Following previous models \citep{Lee2017e2e_corefernce,Zhao2020SpanMlt}, two kinds of scores are generated for the dependency between opinions and aspects: (1) a sentiment mention score determines if a span candidate is an opinion mention; 
	and (2) an aspect-opinion alignment score estimates the modification relation in an aspect-opinion pair.    
	
	\paragraph{Distilled Sentiment Mention Scoring. }
	Most previous span-based alignment models employ an one-dim-out neural module (e.g., MLP with $\sigmoidii$) to determine the confidence of a mention but require full supervision for accurate predictions. Considering such supervision is unavailable in ALSC, we consequently weaken the supervision from ``whether a mention is a gold opinion term'' to ``whether a mention expresses a sentiment polarity''. The weak supervision can be readily obtained from a well-trained phrase-level sentiment classification model via distilling model \citep{Hinton2015distillation}. 
	Hence, we first employ an MLP-based classifier built upon an opinion term candidate $\vu^o$ to derive a three-categorical sentiment distribution, 
	\begin{align}
		\vp^{(ms)}_o &= \softmax(\mlp(\vu^o; \theta^{(ms)}))\!\in\!\R^{3};
		% \\ s^o^s &= 1 - p^s_1
	\end{align}
	and a phrase-level sentiment classification model is
	\begin{align}
		\bar \vp^{(pl)}_o = \sentimodel(\vo; \theta^{(pl)}),
	\end{align}
	which is also based on a pre-trained Transformer and trained on a popular phrase-level sentiment analysis dataset, Stanford Sentiment Treebank \citep{Socher2013SST}. Then, we can define a soft loss of sentiment distillation based on Kullback–Leibler (KL) divergence, i.e.,
	\begin{align}
		\gL^{(kl)} = \kl(\bar\vp^{(pl)}_o|| \vp^{(ms)}_o). 
	\end{align}
	Last, we use the non-neutral probability as the confidence of a mention expressing sentiment polarity, i.e., the sentiment mention score, 
	\begin{align}
		r^o &\coloneqq 1 - \vp^{(ms)}_o {[\hat y = \text{Neutral}]},~\forall\vo\in\gO, \\
		\vr &= \{r^o\}_{\vo\in\gO}\in\R^m. \label{eq:sentiment_score}
	\end{align}

	\paragraph{Aspect-Opinion Dependency Modeling. } 
	Besides the mention score, we use an MLP to estimate the alignment between the aspect and opinion candidates. First, we obtain a relationship representation by an interactive concatenation \citep{Reimers2019Sentence_BERT} of their span-level embeddings,% i.e., 
	\begin{align}
		\vq^o = [\vc; \vu^o; \vc \odot \vu^o; \vz^o],~\forall \vo\in\gO,
	\end{align}
	where ``$\odot$'' is Hadamard product and $\vz$ is a learnable relative-position embedding indicating their distance over the syntactic dependency parsing tree of the sentence. 
	Again, please note we omitted the superscript $a$ in the equation for simplification. 
	Next, $\vq^o$ is passed into an MLP-based scorer to calculate pairwise alignment score, i.e.,
	\begin{align}
		\tilde f^o &= \mlp(\vq^o; \theta^{(as)}) \in\R,~\forall\vo\in\gO, \\
		\tilde \vf &= \{f^o\}_{\vo\in\gO}\in\R^m.
	\end{align}
	Then, we apply $\softmax$ to $\tilde \vf$ for normalized alignment scores, which are subsequently weighted by the corresponding sentiment mention scores $\vr$ in Eq.(\ref{eq:sentiment_score}) to derive the dependency scores between the aspect and all opinion candidates, i.e., 
	\begin{align}
		\vf = \softmax(\tilde \vf) \odot \vr. \label{eq:final_score}
	\end{align}
	
	\paragraph{``Dummy'' Opinion Term. } However, a special scenario has been often ignored when capturing the aspect-opinion dependency or alignment. That is, an aspect term may not correspond to any opinion (statistically, $>65\%$ of neutral-labeled aspects in benchmarks w/o opinion).
	To remedy this, we adopt a concept of ``\textit{dummy span}'' that indicates no opinion term in the sentence towards an aspect. 
	We can imply aspect-``dummy'' dependency score by considering the neutral-sentiment probability of the aligned opinion term candidates, i.e., 
	\begin{align}
		f^{(d)}&= \delta \sum (\softmax(\tilde \vf) \odot (\bm{1} - \vr)), \label{eq:dummy_score}
	\end{align}
	where $\delta$ denotes a re-scaling hyperparameter. So we rewrite the normalized dependency scores as 
	\begin{align}
		\vf \leftarrow [\vf; f^{(d)}] \in\R^{m+1}.
	\end{align}

	It is also essential to integrate the aligned opinion term candidates (including the dummy opinion) into the sentiment-agnostic aspect representation $\vc$. 
	Specifically, we first obtain opinion representation by an attention-like operation, i.e.,
	\begin{align}
		\vu = [\mU, \vc] \cdot \vf,
	\end{align}
	where ``$\cdot$'' denotes matrix multiplication and, as following \citet{Lee2017e2e_corefernce}, the aspect representation itself corresponds to the dummy opinion, i.e., $[\mU, \vc]\in\R^{d\times (m+1)}$. Lastly, we integrate the opinion representation to the aspect one by gating, 
	\begin{align}
		\vg &= \sigmoidii(\mlp([\vc;\vu];\theta^{(gm)})) \in\R^{3d}, \\
		\vv &= \vg \cdot \vc + (\1 - \vg) \cdot \vu.
	\end{align}
	As a result, $\vv$ stands for opinion-enrich aspect representation and is ready for sentiment classification. 
	
	\begin{table}[t]\small 
\renewcommand\tabcolsep{4.0pt}
	\centering
	\begin{tabular}{ccccccc}
	\hline
    \multirow{2}{*}{Dataset} & \multicolumn{2}{c}{Positive} & \multicolumn{2}{c}{Neutral} & \multicolumn{2}{c}{Negative}  \\ \cline{2-3} \cline{4-5} \cline{6-7}
                             & Train & Test                 & Train & Test                & Train & Test                  \\ \hline
    Laptop14                 & 994   & 341                  & 464   & 169                 & 870   & 128                     \\
    Rest14                   & 2164  & 728                  & 637   & 196                 & 807   & 196                   \\
    Rest15                   & 912   & 326                  & 36   & 34                   & 256   & 182                   \\
    Rest16                   & 1240  & 469                  & 69   & 30                   & 439   & 117                   \\    
    Twitter                  & 1561  & 173                  & 3127  & 346                 & 1560  & 173                   \\ \hline
    \end{tabular}
	\caption{\small Summary statistics of five benchmark datasets. }
	\label{tb:benchmark_alsc_stat}
\end{table}
	\subsection{Model Training and Inference} \label{sec:train-infer}
	
	\paragraph{Aspect-level Sentiment Classification.} On top of $\vv$, we define a neural classifier for the final three-categorical sentiment prediction as
	\begin{align}
		\vp^{(sc)} = \softmax(\mlp(\vv ;\theta^{(sc)})) \in\R^3.
	\end{align}
	And the training loss of ALSC task is written as
	\begin{align}
		\gL^{(sc)} = - \sum\nolimits_\gD \sum\nolimits_{\gA} \log \vp^{(sc)}_{[\hat y = y^{(sc)}]}, 
	\end{align}
	where $y^{(sc)}$ denotes the oracle label of an asepct.

	\paragraph{Training and Inference. } Besides the discriminator loss $\gL^{(dis)}$ in Eq.(\ref{eq:discriminator_loss}), we train the learnable parameters in our proposed SARL model towards a linear combination of the other three losses, i.e., 
	\begin{align}
		\gL^{(alsc)} = \gL^{(sc)} + \beta\gL^{(adv)} + \gamma\gL^{(kl)}. \label{eq:alsc_loss}
	\end{align}
	We also set a hyper-parameter $\alpha$ to control the proportion of discriminator learning (i.e., $\gL^{(dis)}$) against the ALSC model learning (i.e., $\gL^{(alsc)}$). The inference procedure can be simply written as
	\begin{align}
		y^* = \arg\max\vp^{(sc)}. 
	\end{align}

	\paragraph{Unsupervised Opinion Extraction. }
	A well-trained SARL model is equipped with the capability to extract opinion term(s) for an aspect, based on its intermediate variable, i.e.,  
	\begin{align}
		\vo^* = \arg\max\nolimits_{ \{o,\text{dummy}\} } \vf,
	\end{align}
	where $\vf\in\R^{m+1}$ from Eq.(\ref{eq:final_score}) denotes aspect-opinion dependency scores, including the last dim for dummy opinion, i.e., an aspect w/o any opinion in the sentence. 
	Thereby, $\vo^*$ is the extracted opinion term for a specified aspect term in the sentence, and it is worth mentioning again that the opinion extraction is learned in an unsupervised manner.

	\section{Experiment}
	
	\begin{table}[t]\small 
\renewcommand\tabcolsep{2pt}
	\centering
	\begin{tabular}{cccc|cccc}
	\hline
    Dataset   & \#AT   & \#OT  & \#$\theta$  & Dataset   & \#AT  & \#OT  & \#$a^{neu/o}$ \\ \hline
    Laptop14  & 444  & 527   & 112    & Rest14    & 832   & 974  & 137 \\ \hline
    \end{tabular}
	\caption{\small Summary statistics of unsupervised opinion extraction test set. 
	\#$a^{neu/o}$ is the number of neutral-sentiment aspects without corresponding opinion terms. 
	}
	\label{tb:benchmark_inter_stat}
\end{table}
	
	\begin{table}[t] \small
\renewcommand\tabcolsep{1pt}
    \centering
    \begin{tabular}{ccccccc} \hline
        LR ($\times 10^{-5}$) & size & Laptop14 & Rest14 & Rest15 & Rest16 & Twitter \\ \hline
        \multirow{2}{*}{lr-$\theta^{(ptm)}$} & base      & 1.3 & 2   & 2   & 2   & 2 \\
                                    & large     & 1   & 1   & 1   & 1   & 2 \\ \hline
        \multirow{2}{*}{lr-others}  & base     & 20  & 1   & 20  & 20  & 20 \\ 
                                    & large     & 20  & 1   & 1   & 20  & 1 \\ \hline
    \end{tabular}
    \caption{\small Settings of learning rate. ``others'' includes $\theta^{(aa)}$, $\theta^{(dis)}$, $\theta^{(ao)}$, $\theta^{(ms)}$, $\theta^{(as)}$, $\theta^{(gm)}$ and $\theta^{(sc)}$.}
    \label{tb:hyperparams}
\end{table}
	
	\begin{table*}[t] \small
\renewcommand\tabcolsep{4.0pt}
	\centering
	\begin{tabular}{clcccccccccc}
		\hline
		 \multirow{2}{*}{\textbf{Embedding}} & \multirow{2}{*}{\textbf{Method}}   & \multicolumn{2}{c}{\textbf{Laptop14}}  & \multicolumn{2}{c}{\textbf{Rest14}}   & \multicolumn{2}{c}{\textbf{Rest15}}   &\multicolumn{2}{c}{\textbf{Rest16}}  & \multicolumn{2}{c}{\textbf{Twitter}} \\ \cline{3-12}
		 & & Accu     & Ma-F1          & Accu     & Ma-F1           & Accu     & Ma-F1   & Accu     & Ma-F1  & Accu     & Ma-F1\\ \hline
		 \multirow{2}{*}{Static} 
		 &ASGCN              & 75.55    & 71.01        & 80.86    & 72.19    & 79.89 & 61.89     & 88.99 & 67.48  & 72.15    & 70.40          \\
		 \multirow{2}{*}{Embedding} 
		 &CDT                & 77.19    & 72.99        & 82.30    & 74.02    & - & -             & 85.58 & 69.93      & 74.66    & 73.66          \\
		 &BiGCN 		     & 74.59	  & 71.84        & 81.97    & 73.48	   & 81.16 & 64.79     & 88.96 & 70.84   & 74.16	  & 73.35          \\ \hline
		 
		 \multirow{3}{*}{BERT$_\text{base}$} 
 &DGEDT     & 79.80    & 75.60        & 86.30    & 80.00    & 84.00 & 71.00     & 91.90 & 79.00 & 77.90    & 75.40           \\
  &RGAT      & 78.21    & 74.07        & 86.60    & 81.35    & - & -             & - & -       & 76.15    & 74.88            \\ 
 &kumaGCN    & 81.98    & 78.81        & 86.43    & 80.30    & 80.69 & 65.99     & 89.39 & 73.19  & 77.89 & 77.03    \\ \hline
        \multirow{5}{*}{RoBERTa$_\text{base}$} 
        &$\dag$ASGCN-FT-RoBERTa              & 83.33    & 80.32        & 86.87    & 80.59    & - & -            & - & -          & 76.10 & 75.07      \\
        &$\dag$PWCN-FT-RoBERTa               & 84.01    & 81.08        & 87.35    & 80.85    & - & -            & - & -          & 77.02 & 75.52          \\
        &$\dag$RGAT-FT-RoBERTa               & 83.33    & 79.95        & 87.52    & 81.29    & - & -            & - & -          & 75.81 & 74.91             \\ 
        &$\dag$MLP                           & 83.78    & 80.73        &87.37     & 80.96    & - &-             & - & -          & 77.17 & 76.20 \\\cline{2-12}
        &\textbf{SARL} (ours)                         & 85.42    & \textbf{82.97}        & 88.21 & 82.44       & 88.19 & 73.83    & 94.62 & 81.92  & 78.03 & 76.97                   \\ \hline
        RoBERTa$_\text{large}$
        &\textbf{SARL} (ours)                & \textbf{85.74}    & \textbf{82.97}       & \textbf{90.45} & \textbf{85.34}        & \textbf{91.88} & \textbf{78.88}     & \textbf{95.76} & \textbf{84.29}  & \textbf{78.32} & \textbf{77.32}              \\ \hline
	\end{tabular}
	\caption{\small 
	ALSC results on the five datasets. $\dag$Numbers are from \citet{Dai2021does-syntax}, and others are from the original papers, i.e., ASGCN \citep{Zhang2019ASGCN}, CDT \citep{Sun2019CDT}, BiGCN \citep{Zhang2020BiGCN}, DGEDT \citep{Tang2020DGEDT}, RGAT \citep{wang2020RGAT}, kumaGCN \citep{chen2020kumaGCN}, PWCN-FT-RoBERTa \citep{Zhang2019Syntax-aware}. 
	}
	\label{tb:main_results}
\end{table*}

	% \subsection{Experiment Settings}

	\paragraph{Datasets.}
	
	For ALSC task, we evaluate our model on five datasets\footnote{The source code and datasets are available at \url{https://github.com/wangbo9719/SARL_ABSA}}, whose statistics are listed in Table~\ref{tb:benchmark_alsc_stat}, 
	including (i) Laptop14 (SemEval-2014T4) \citep{Pontiki2014SemEval14} with laptop reviews, (ii) Rest14 (SemEval-2014T4), Rest15 (SemEval-2015T12) and Rest16 (SemEval-2016T5)  \citep{Pontiki2014SemEval14,PontikiG2015rest15,Pontikl2016Rest-16} with restaurant reviews, and (iii) Twitter \citep{Mitchell2013Twitter_set} with tweets. 
	Following most competitors including the models compared in Table \ref{tb:main_results}, we do not split training set.  
	
	To evaluate unsupervised opinion extraction, we employ the test set from \citet{Xu2020Position-aware} where annotations of opinion terms\footnote{The sentences in this dataset cannot completely match the standard ALSC dataset so we only use the overlap part.} are from \citet{Fan2019Target-oriented}. The statistics are listed in Table~\ref{tb:benchmark_inter_stat}.

	\begin{table*}[t] \small
\renewcommand\tabcolsep{2.5pt}
    \centering
    \begin{tabular}{c|l|cccc|cccc|cccc|cccc} \hline
         \multirow{2}{*}{\textbf{Super-}} &\multirow{3}{*}{\textbf{Method}} & \multicolumn{8}{c|}{\textbf{Laptop14}}  & \multicolumn{8}{c}{\textbf{Rest14}}  \\ \cline{3-18}
         \multirow{2}{*}{\textbf{vision}} & & \multicolumn{4}{c|}{@1}   & \multicolumn{4}{c|}{@3}  & \multicolumn{4}{c|}{@1}    & \multicolumn{4}{c}{@3}  \\ \cline{3-18}                   
		 & &EM &P &R &F      &EM &P &R &F   &EM &P &R &F   &EM &P &R &F    \\ \hline
		 \multirow{2}{*}{\checkmark}   &SpanMlt     &- &- &- &80.6 &- &- &- &-              &-&-&-&84.0 &-&-&-&- \\
		 &\citet{Peng2020Knowing} & - & 81.8 & 84.8 & 83.2 &- &- &- &-                           & - & 76.9 & 75.3 & 76.0 &-&-&-&- \\ \hline
		 \multirow{4}{*}{\XSolidBrush} &Induced Tree & 8.5 & 29.4 & 20.6 & 22.6 & - & - & - & -   & 10.6 & 30.2 &21.8 & 23.5 & - & - & - & -  \\ 
		 &\textbf{SARL} (ours) &\textbf{42.3} &\textbf{62.2} &\textbf{68.7} & \textbf{62.8} & \textbf{60.5} &\textbf{76.3} & \textbf{81.4}  & \textbf{77.1}     & \textbf{49.4} &\textbf{65.9} &\textbf{67.1} &\textbf{64.8}    & \textbf{76.0} &\textbf{86.0} &\textbf{88.2} &\textbf{86.1}        \\ \cline{2-18}   
         &~~w/o distillation  &16.3 &37.5 &32.1 &31.8 &27.5  &52.8  &50.7 &47.8   &19.0 &43.9 &45.3 &41.1   &34.8 &65.3 &68.1 &62.8               \\  
 &~~w/o sentiment score   &15.0 &32.7 &29.7 &28.6   &27.7 &54.2 &51.9 &49.4         &30.9 &52.5 &50.5 &48.9  &52.1 &73.7  & 74.1  & 71.2            \\ \hline
    \end{tabular}
    \caption{\small Unsupervised opinion extraction results (\%) on Laptop14/Rest14. 
    % The numbers here are percentages. 
    The resulting numbers of SpanMlt \citep{Zhao2020SpanMlt} and \citet{Peng2020Knowing} are from their original papers which the metrics under @1 are the standard metrics. 
    The results of induced tree \citep{Dai2021does-syntax} are calculated by regarding the aspects' sub-nodes in the tree as its extracted opinions. 
    }
    \label{tb:interpretable_resutls}
\end{table*}
	
	\paragraph{Training Setups.}
	We use a mini-batch Stochastic Gradient Descent (SGD) to minimize the loss functions, with  Adam optimizer, $10\%$ warm-up, and a linear decay of the learning rate.
	To initialize the Transformer, we alternate between pre-trained RoBERTa$_\text{base}$ and $_\text{large}$. 
	We set batch size $=16$ and max sequence length $=64$ based on experience, and conduct grid searches for the other hyperparameters. 
	Then, we set the max width of candidate span $l=15$, the number of training epochs $=7$ for Twitter and $10$ for other datasets, the training proportion $\alpha=1/3$ for Laptop14, $0.2$ for Rest14 and Rest15, $0.1$ for Rest16, the loss weight in Eq.(\ref{eq:alsc_loss}) $\beta=0.05$ for Laptop14 and Rest16 and $0.1$ for the remains, $\gamma=1$, and $\delta=1/m$ in Eq.(\ref{eq:dummy_score}). 
	The learning rates are listed in Table \ref{tb:hyperparams}.

	% Results and Analysis
	
	\subsection{Overall Performance}
	ALSC results of competitive approaches and our SARL on the five benchmarks are shown in Table \ref{tb:main_results}. 
	Following prior works, we adopt accuracy (Accu) and macro-F1 (Ma-F1) to evaluate the performance, and the results of SARL are the best values from ten runs. 
	It is observed that our proposed SARL achieves state-of-the-art performance on all these datasets. 
	Compared to static embedding-based methods, the methods with pre-trained Transformer gain better results. 
	In the same embedding genre RoBERTa$_{\text{base}}$, SARL outperforms others by an average of 1\% on accuracy and 1.3\% on macro-F1. 
	Furthermore, the RoBERTa$_{\text{large}}$-based SARL derives more significant progress.

	\subsection{Sentiment Bias Elimination}
	
	\begin{figure}[b]
		\subfigure{\includegraphics[width=0.17\textwidth]{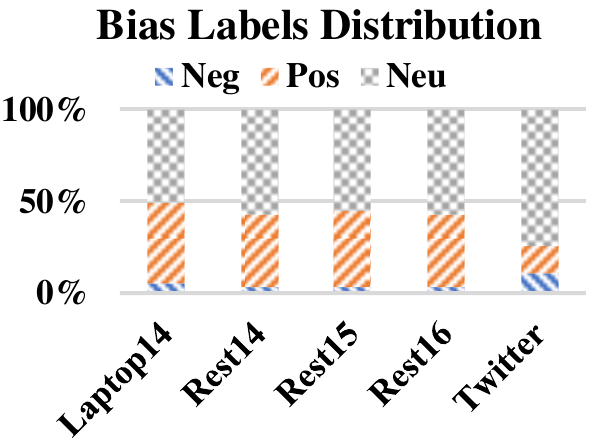}}
		\subfigure{\includegraphics[width=0.3\textwidth]{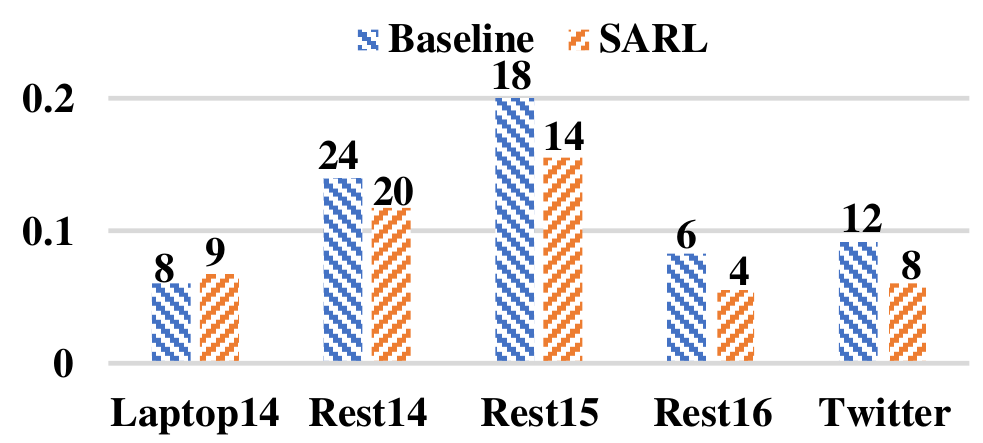}}
		\vspace{-3mm}
		\caption{\small Sentiment bias analysis on five datasets. The ``Baseline'' in right denotes the span-based Transformer baseline. Suppose an aspects set $\gS=\{a|(y^{(pr)} \neq \textit{Neutral}) \wedge (y^{(pr)} \neq y^{(sc)})\}$ denotes all potential aspects who will be misclassified due to the sentiment bias, the vertical axis in right represents the proportion of ${|\gQ|}/{|\gS|}$, where $\gQ = \{a|a \in \gS \wedge y^* = y^{(pr)} \neq y^{(sc)}\}$. And the number on the top of bin is $|\gQ|$. 
		}
		\label{fig:sentiment_bias} 
	\end{figure}
	\begin{figure}
		\centering
		\includegraphics[width=0.45\textwidth]{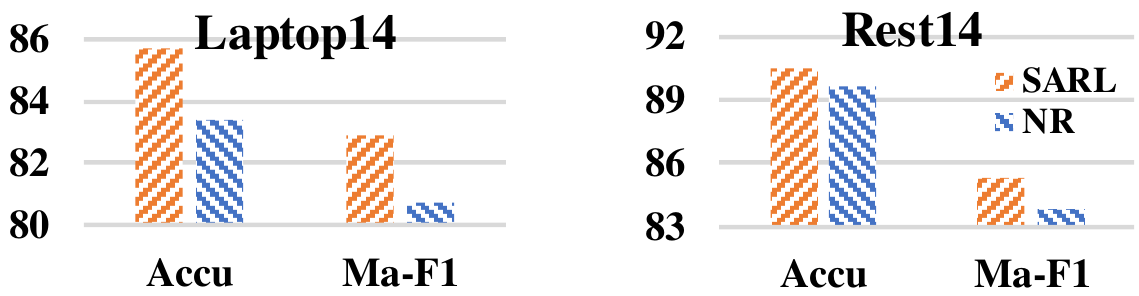}
		\vspace{-2mm}
		\caption{\small ALSC results of SARL, and ``SARL with neutral reinforce'', i.e., w/ $\gL^{(adv)}_{\theta^{(dis)}, \theta^{(aa)}, \theta^{(ptm)}}$ while w/o $\gL^{(dis)}$. 
		}
		\label{fig:adv_illustration}
	\end{figure}

	\begin{table}[b] \small
		\centering
		\begin{tabular}{cc|cc} \hline
			\multicolumn{2}{c|}{\textbf{Laptop14}}  & \multicolumn{2}{c}{\textbf{Rest14}}  \\ \hline
			Hits@1 & Hits@3    & Hits@1 & Hits@3 \\ \hline
			30.36 & 56.25 & 45.99 & 61.31 \\ \hline
			
		\end{tabular}
		\caption{\small The Hits@N results about dummy opinions. }
		\label{tb:dummy_results}
	\end{table}

	\paragraph{Bias Statistics.}
	As in Figure \ref{fig:sentiment_bias} (left), the sentiment bias is common since aspects with bias polarity account for nearly 50\% in most datasets.

	\paragraph{The Success of Adversarial Learning.}
	To explore the effectiveness of our adversarial learning, we first train a model without adversarial, ``w/o adv'' for short (please refer to below ablation study for more). 
	Then, based on this trained model, we add a sentiment bias detector that has the same architecture as the discriminator to detect the bias existing in aspects representations. 
	Here, we use the ratio of neutral predictions to measure the accuracy of detector since the smaller ratio means more bias detected. 
	Compared to  98.75\% by full SARL model, the detector's ratio 69.64\% is much smaller, which proves both necessity and success of adversarial learning. 
	Further, as in Figure \ref{fig:adv_illustration}, we train a model w/o adv but with neutral reinforce and there are obvious decreases compared with the full SARL.

	\begin{table}[b] \small
		\renewcommand\tabcolsep{4.0pt}
		\centering
		\begin{tabular}{lcccc}
			\hline
			\multirow{2}{*}{\textbf{Method}}  & \multicolumn{2}{c}{\textbf{Laptop14}}  & \multicolumn{2}{c}{\textbf{Rest14}}   \\ \cline{2-5}
			& Accu     & Ma-F1       & Accu     & Ma-F1 \\ \hline
			Full model                      & 85.74    & 82.97       & 90.45    & 85.34      \\ \hline
			~~w/o adv                 & 83.70    & 80.55       & 89.46    & 82.16 \\
			~~w/o AT-OT aligner               & 84.64    & 81.75       & 89.28    & 83.51  \\
			~~w/o all          & 83.86    & 80.53       & 89.02    & 82.59 \\ 
			\hline
			
		\end{tabular}
		\caption{\small 
			Ablation results of ALSC. The full model denotes our proposed SARL. 
			``w/o adv'' removes anti-bias module in\S\ref{sec:adv_encoder}.
			``w/o AT-OT aligner'' removes the proposed module in \S\ref{sec:asp-op_dep}.
			``w/o all'' degrades our SARL model to Transformer-based span-level ALSC (i.e., classification based on Eq.(\ref{eq:span_asp_emb})). 
		}
		\label{tb:ablation_results}
	\end{table}
	
	\paragraph{Results of Elimination.}
	Comparing to a span-level baseline that simply feeds the aspect representation $\vc$ in Eq.(\ref{eq:span_asp_emb}) into an MLP to get the sentiment predictions, ours alleviates the problem as shown in Figure \ref{fig:sentiment_bias} (right). A possible reason is that the aspects in Laptop14 (e.g., \textit{screen} and \textit{size}) merely have a little sentiment in themselves so that the bias labels from SentiWordNet are noisier. 
	
	\begin{table*}[t] \small
		\centering
		\setlength{\tabcolsep}{5pt}
		\begin{tabular}{clccll}
			\hline
			\textbf{Index} &\textbf{Example}      & \textbf{``w/o all''}   & \textbf{SARL}  & \textbf{Top-3 candidates} \\ \hline
			
			\multirow{2}{*}{1} &After dinner I heard [music]$_{neu}$ playing and   & \multirow{2}{*}{\textit{pos}} & \multirow{2}{*}{\textit{neu}} & dummy; playing;  \\   
			&discovered that there is a lounge downstairs.       & & &  playing and discovered \\ \hline
			2 &Desserts include [flan]$_{neu}$ and sopaipillas.      & \textit{pos} & \textit{neu} & dummy; and; include \\ \hline
			
			3 &How is this [place]$_{neg}$ still open?    & \textit{pos} & \textit{neg} & open; still open; dummy \\  \hline \hline 
			\multirow{2}{*}{4} &This place has beautiful [sushi]$_{pos}$,    & \multirow{2}{*}{\textit{pos}} & \multirow{2}{*}{\textit{pos}} & delicious; delicious CHEAP;  \\ 
			&and it's delicious CHEAP.   & & &  beautiful \\ \hline
			
			\multirow{2}{*}{5}   &\multirow{2}{*}{The wait [staff]$_{pos}$ was loud and inconsiderate.}     & \multirow{2}{*}{\textit{neg}} & \multirow{2}{*}{\textit{neg}} & loud; was loud;  \\ 
			&  &  & & loud and inconsiderate \\ \hline \hline
			
			\multirow{2}{*}{6}  &The sauce is excellent (very fresh) with   & \multirow{2}{*}{\textit{pos}} & \multirow{2}{*}{\textit{pos}}  &  excellent; excellent (very fresh); \\
			&$\text{[dabs of real mozzarella]}_{neu}$.  & &                                                              &  fresh\\ \hline
			
			7 &15\% gratuity automatically added to the [bill]$_{neg}$.  & \textit{neu} & \textit{neu} &dummy; automatically; 15 \\ \hline
			
			\multirow{2}{*}{8}  &The [quality of the meat]$_{neg}$ was on par & \multirow{2}{*}{\textit{pos}} & \multirow{2}{*}{\textit{neu}} & \multirow{2}{*}{dummy; par; was on par}\\
			&with your local grocery store. & & & \\ \hline

		\end{tabular}
		\caption{\small Case study for sentiment bias (row 1-3), unsupervised opinion extraction (row 4-5) and error analysis (6-8). 
		}
		\label{tb:analysis_sentiment_bias}
	\end{table*}

	\subsection{Unsupervised Opinion Extraction}
	\paragraph{Explicit Opinion Extraction.}
	To measure the model's ability of opinions extraction w/o supervision, we present the following novel metrics including the top-N-based Exact Match (EM@N), Precision (P@N), Recall (R@N) and F1-Score (F@N). 
	Specifically, EM@N denotes the gold opinion term appears in the top-N opinion term candidates. 
	And the Precision@N, Recall@N and F1@N are also employed to describe the maximum char-level overlap. 
	As shown in Table \ref{tb:interpretable_resutls}, SARL achieves promising performance on unsupervised opinion extraction. 
	Without any opinion annotation data, the top-3 based metrics achieve similar or even better performance compared with the models under opinion supervision. 
	Further, comparing with Induced Tree, SARL achieves far better performance.

	\paragraph{Dummy Opinion Extraction.}
	To measure the dummy opinion ranking about neutral-sentiment aspects without opinion terms, we apply Hits@N that stands for the ratio of such aspects that the dummy opinion is ranked in top-N. 
	As shown in Table \ref{tb:dummy_results}, the performance is adequate to support the purpose of introducing dummy opinions. 
	
	% Ablation
	\subsection{Ablation Study}
	
	To explore each module's contribution, we conduct an extensive ablation study. 
	For ALSC, 
	(1) The results of ``model w/o adv'' are slightly higher than the ``model w/o all'', which indicates that the AT-OT aligner is helpful for classification due to our opinion terms integration; 
	and (2) The ``model w/o aligner'' achieves sub-optimal results and completely loses the ability of opinion extraction. 
	In addition, the component dropping also severely affects the performance of opinion extraction as in Table \ref{tb:interpretable_resutls}. 
	The extraction performance decrease of SARL ``w/o distillation'' is larger than ``w/o sentiment score''. 
	A potential reason is that the sentiment scorer performs poorly without any supervision and thus generates incorrect sentiment scores.

	\subsection{Case Study}
	\paragraph{Does SARL eliminate sentiment bias?} % [table1]
	As the top-3 rows in Table \ref{tb:analysis_sentiment_bias}, those aspects with intrinsic positive bias are mis-classified by our span-level baseline but correctly classified by SARL, which verifies the effectiveness of SARL.

	\paragraph{How does SARL obtain interpretability from unsupervised opinion extraction?} 
	As listed in Table \ref{tb:analysis_sentiment_bias}, SARL can exactly extract the opinion terms of the targeted aspect so provide promising interpretability for sentiment prediction under the unsupervised opinion setting. 
	In addition, for the neutral aspect without explicit opinion terms in a sentence, the dummy opinion always ranks first like the $2^{nd}$ row in Table \ref{tb:analysis_sentiment_bias}, which explains the reason for neutral prediction. 
	
	\paragraph{Error Analysis.} 
	To analyze the limitation of ALSC models including ours, we investigate all the examples mis-classified by SARL on Rest14, and summarize two main problems: (1) The major (up to 66\%) problem is neutral-related mis-classifying because a neutral aspect term is affected by the polarity words associated with other aspects (e.g., $6^{rd}$ row). 
	The other problem ($23\%$) refers to that it is infeasible to determine aspects' sentiment without commonsense knowledge (e.g., $7^\text{th}$ row) or additional information (e.g., $8^\text{th}$ row). 
	In summary, how to accurately classify the aspect without explicit opinion is still an open problem.

	\section{Related Work}

	\paragraph{Aspect-Level Sentiment Classification.}
	
	ALSC relies heavily on modification relations between aspect term and opinion term in a sentence, so recent progresses mainly fall into modeling the relations by applying graph neutral networks to dependency parsing tree \citep{Sun2019CDT,Zhang2019ASGCN,wang2020RGAT}. Despite their effectiveness, they lack interpretability to the sentiment prediction. 
	For further boosted, many methods \citep{sun2019bert_qa,hu2019SpanABSA, chen2020kumaGCN, Mao2021Dual-MRC, Dai2021does-syntax} introduce the pre-trained Transformer, which also brings a little interpretability due to the highlight modification between aspect and opinion terms derived from potential syntax knowledge \citep{Wu2020Perturbed-masking,Dai2021does-syntax}. However, the derived interpretability is far from human-level. 
	Furthermore, the Transformer tends to internalize terms' intrinsic sentiment bias, which is harmful to ALSC. 
	
	\paragraph{Aspect-Opinion Alignment.}
	\citet{Fan2019Target-oriented} first defined the aspect-oriented opinion extraction task in which the aspect terms are given in advance. 
	Later, \citet{Zhao2020SpanMlt} proposed the aspect-opinion co-extraction task. 
	More recently, \citet{Peng2020Knowing} proposed the aspect sentiment triplet extraction and further explorations have conducted by \citet{Xu2020Position-aware} and \citet{Mao2021Dual-MRC}. 
	These models extract the opinion terms for aspects and thus provide explicit interpretability. However, they require manually labeled opinions data for training, which is much more expensive than three-categorical labeling in ALSC.

	\section{Conclusion}
	In this work, we propose SARL framework for ALSC. Specifically, we first present an adversarial anti-bias aspect encoder to eliminate sentiment bias in aspects and then propose an aspect-opinion dependency aligner to unsupervisedly extract opinions. The experiments on 5 benchmarks can greatly support our motivations and empirical results show state-of-the-art performance with interpretability.

	\section*{Acknowledgement}
	This work is supported by the National Natural Science Foundation of China (No.61976102, No.U19A2065).
	
	\bibliography{references}
	\bibliographystyle{acl_natbib}

\end{document}